\documentclass[sigconf]{acmart}
\usepackage{graphicx}
\usepackage{subfigure}
\usepackage{array}
\usepackage{dcolumn}
\usepackage{multirow}
\usepackage{multicol} 
\usepackage{float}
\usepackage{amsmath}
\usepackage{bm}
\usepackage{fancyhdr}
\usepackage{hyperref}
\usepackage[linesnumbered,ruled,lined,boxed,commentsnumbered]{algorithm2e}
\usepackage{makecell} 
\usepackage{mathrsfs} 
\usepackage{balance}
\AtBeginDocument{%
  \providecommand\BibTeX{{%
    \normalfont B\kern-0.5em{\scshape i\kern-0.25em b}\kern-0.8em\TeX}}}

\copyrightyear{2023}
\acmYear{2023}
\setcopyright{acmlicensed}
\acmConference[MM '23] {Proceedings of the 31st ACM International Conference on Multimedia}{October 29--November 3, 2023}{Ottawa, ON, Canada.}
\acmBooktitle{Proceedings of the 31st ACM International Conference on Multimedia (MM '23), October 29--November 3, 2023, Ottawa, ON, Canada}
\acmPrice{15.00}
\acmISBN{979-8-4007-0108-5/23/10}
\acmDOI{10.1145/3581783.3612495}

\begin{document}

\title{Counterfactual Cross-modality Reasoning for Weakly Supervised Video Moment Localization}

\author{Zezhong Lv}
\email{zezhonglv0306@gmail.com}
\affiliation{%
  \institution{Gaoling School of Artificial Intelligence, Renmin University of China \\ Beijing Key Laboratory of Big Data Management and Analysis Methods}
  \city{Beijing}
  \country{China}
}
\author{Bing Su}
\authornote{Corresponding author.}
\email{subingats@gmail.com}
\affiliation{%
  \institution{Gaoling School of Artificial Intelligence, Renmin University of China \\ Beijing Key Laboratory of Big Data Management and Analysis Methods}
  \city{Beijing}
  \country{China}
}
\author{Ji-Rong Wen}
\email{jrwen@ruc.edu.cn }
\affiliation{%
  \institution{Gaoling School of Artificial Intelligence, Renmin University of China \\ Beijing Key Laboratory of Big Data Management and Analysis Methods}
  \city{Beijing}
  \country{China}
}

\begin{abstract}

Video moment localization aims to retrieve the target segment of an untrimmed video according to the natural language query. Weakly supervised methods gains attention recently, as the precise temporal location of the target segment is not always available. However, one of the greatest challenges encountered by the weakly supervised method is implied in the mismatch between the video and language induced by the coarse temporal annotations. To refine the vision-language alignment, recent works contrast the cross-modality similarities driven by reconstructing masked queries between positive and negative video proposals. However, the reconstruction may be influenced by the latent spurious correlation between the unmasked and the masked parts, which distorts the restoring process and further degrades the efficacy of contrastive learning since the masked words are not completely reconstructed from the cross-modality knowledge. In this paper, we discover and mitigate this spurious correlation through a novel proposed counterfactual cross-modality reasoning method. Specifically, we first formulate query reconstruction as an aggregated causal effect of cross-modality and query knowledge. Then by introducing counterfactual cross-modality knowledge into this aggregation, the spurious impact of the unmasked part contributing to the reconstruction is explicitly modeled. Finally, by suppressing the unimodal effect of masked query, we can rectify the reconstructions of video proposals to perform reasonable contrastive learning. Extensive experimental evaluations demonstrate the effectiveness of our proposed method. The code is available at \href{https://github.com/sLdZ0306/CCR}{https://github.com/sLdZ0306/CCR}.

\end{abstract}



\maketitle
\setcopyright{acmcopyright}
\copyrightyear{2023}
\acmYear{2023}
\acmDOI{10.1145/3581783.3612495}

\acmConference[MM '23]{Proceedings of the 31st ACM International Conference on Multimedia}{ October 29--November 14, 2023}{Ottawa, Canada}
\acmBooktitle{
Proceedings of the 31st ACM International Conference on Multimedia (MM '23), October 29--November 14, 2023, Ottawa, Canada}
\acmPrice{15.00}
\acmISBN{979-8-4007-0108-5/23/10}

\vspace{-0.2cm}
\section{Introduction}
\vspace{-0.1cm}
\label{sec:intro}

\begin{figure}[t]
\vspace{0.0cm}
  \centering
  \setlength{\abovecaptionskip}{-0.0cm}
  \setlength{\belowcaptionskip}{-0.0cm}
  \includegraphics[width=1.0\linewidth]{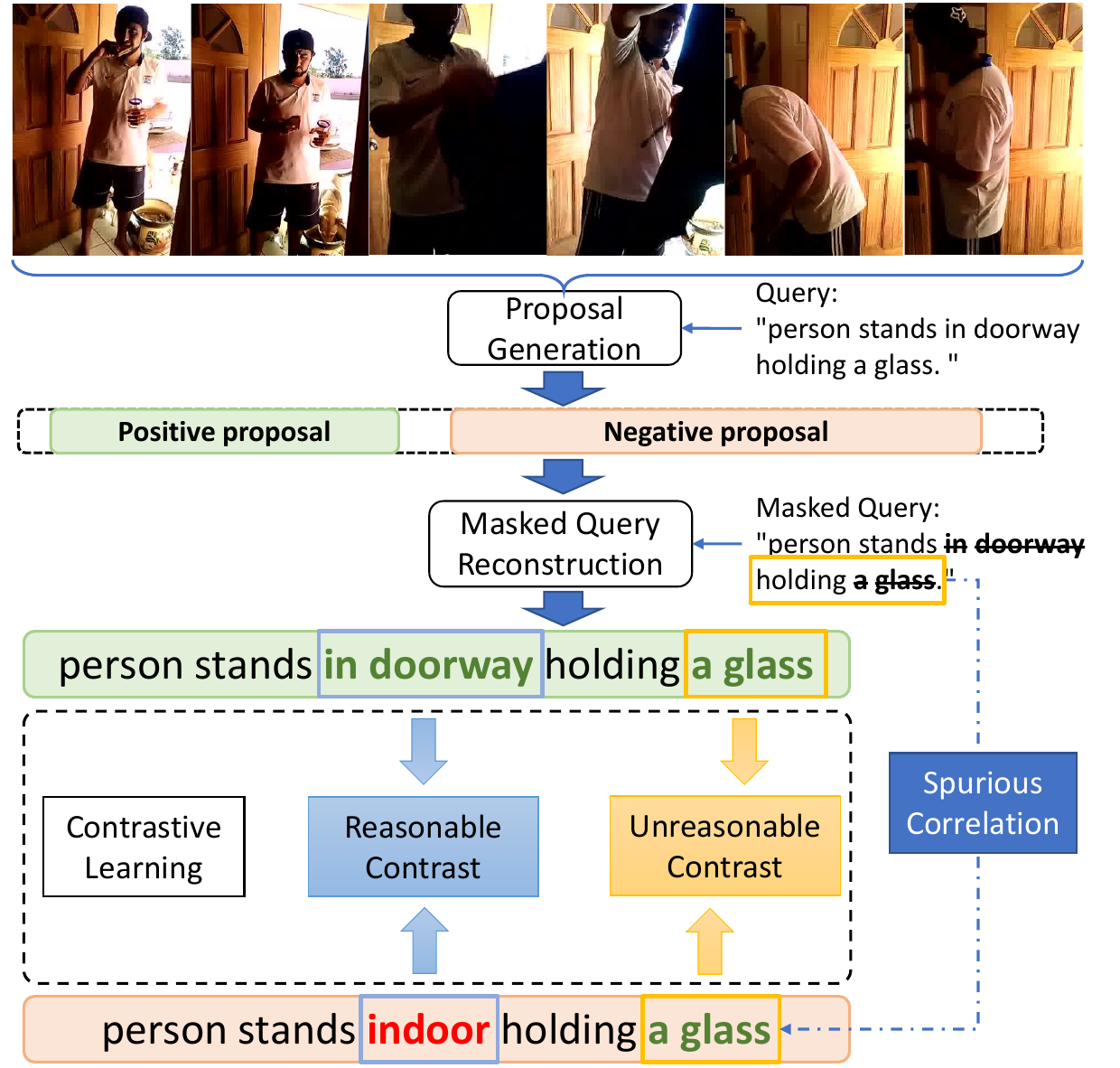}
  \caption{A general query reconstruction based weakly supervised video moment localization framework.
  The reconstruction based on positive and negative proposals is contrasted to learn the cross-modality alignment.
  However, the spurious correlation between un-masked and masked words confuses the contrastive learning process.
  \emph{e.g.} the prediction for the masked ``indoor'' can differ between positive and negative proposals due to changes in the person's location.
  In contrast, the word ``glass'' may still be correctly predicted even if it does not appear in the negative proposal. This abnormal correct prediction is due to a certain pattern learned from the language rather than cross-modality knowledge, which leads to an unreasonable contrast.
  }
  \label{fig:intro}
  \vspace{-0.7cm}
\end{figure}

The widespread use of the internet and mobile devices has led to an unprecedented surge in multimedia content consumption, especially videos, due to their stronger capacity for information expression compared to other media forms. As a result, video understanding has become a crucial challenge in computer vision, encompassing a vast range of research topics, such as video highlight detection \cite{guo2021taohighlight, ye2021temporal} and temporal action localization \cite{gao2017turn, chao2018rethinking, lin2021learning}. While these tasks primarily focus on single video modality, video moment localization \cite{gao2017tall, hendricks2017localizing} is increasingly becoming a central research topic in modeling the relationship between video and natural language, which is largely motivated by the requirements of the cross-modality application scenarios.
Concretely, given a natural language query, video moment localization aims at localizing the start and end boundaries of the target video segment from an untrimmed video according to the semantic contents of the query.
Fully supervised video moment localization methods have achieved promising retrieval performance, which are trained with the exact temporal labels (\emph{i.e.} the start and end times of the target moment) provided in the given datasets.
However, these fine-grained temporal labels are not always accessible in many application scenarios, and manually annotating such labels are expensive and time-consuming.
Besides, the performance of a fully supervised method heavily relies on the quality of the ground truth labels.
Nevertheless, it has been reported that there is a significant man-made temporal bias existing in the distributions of the boundary labels in the widely used benchmarks of video moment localization \cite{lan2023survey}.

Weakly supervised methods are recently \cite{cnm, zhang2020counterfactual, zhang2020regularized} proposed to fix these issues, which are trained with the matched pairs of a whole untrimmed video and its language query and do not require the temporal labels during training anymore.
Due to the lacking of concrete temporal labels, there are no indications of the precise semantic alignment between the query and its corresponding segment in the given video. Therefore, weakly supervised methods can only learn the relationship between vision and language modalities from a more coarse perspective, \emph{i.e.} video-query pairs.
Because there is no fine-grained vision-language alignment provided in the training process, the cross-modality interactions can be confused by the latent mismatch caused by the coarse relationship.
One way to build the connection from such coarsely matched pairs is the mask-reconstruction technique \cite{lin2020weakly}, which partially masks one sample and learns to reconstruct the masked part from the other sample.
This methodology requires no fine-grained annotations to find and model the corresponding information shared between the two modalities.
Concretely, recent state-of-the-art weakly supervised video moment localization methods \cite{cnm,cpl} apply this technique through learning a cross-modality fusion model to predict the masked query based on the video moment candidates, and using the accuracy of the prediction as the measurement of the similarity to apply contrastive learning.

As illustrated in Figure \ref{fig:intro}, one of the common end-to-end schemes utilized by existing weakly supervised methods can be summarized as follows:
(1) Generating several candidate proposals from the untrimmed video based on the cross-modality features, and extracting their features;
(2) By interacting the proposal features with the masked query, the reconstructed query for each proposal is obtained, while the reconstruction loss is taken as the proxy of the alignment score between video proposal and query;
These two steps enable the model to discover the fine-grained cross-modality alignment with no temporal labels;
(3) Applying intra-sample contrastive learning between the reconstruction errors of the positive and negative proposals to train the model.
By comparing the reconstruction-driven alignment scores obtained by the positive and negative proposals respectively, the model is forced to recognize the difference between the well-aligned video segment and the irrelevant ones with respect to the given query, and thus generates promising proposals as the final localization results during inference.
The core idea of these methods relies on that a larger similarity between the video proposal and the query implies the masked query can be reconstructed easier according to the video proposal, and vice versa.

However, when training the cross-modality interaction module on biased queries in the dataset, which illustrate certain patterns of word combinations, the model learns the spurious correlations implied by these combinations.
For example, the non-uniform joint distribution of the noun and predicate in queries reported in \cite{yoon2022selective} will make the model tend to predict the masked word based on the combination of regular word pairs.
As illustrated in Figure \ref{fig:intro}, given a query ``person stands in doorway holding a glass'', when we mask the word ``glass'' and reconstruct it, the answer could be predicted directly based on the language knowledge rather than the cross-modality fusion model because of the highly statistic correlation between the word ``glass'' and ``holding'' implying by dataset.
In other words, the difficulty of the query reconstruction is significantly lower because the masked words can be approximately reconstructed even if the mismatching video moment proposal is fed into the cross-modality model.
Thus, these spurious correlations will distort the similarity measurement generated by the cross-modality model, which is the fundamental of this paradigm to correlate vision and language information in semantic space, and further degrade the performance of contrastive learning.

We propose to discover and mitigate these spurious correlations through Counterfactual Cross-modality Reasoning (CCR).
Specifically, we disentangle the total causal effect on the reconstruction of the masked query as an aggregation of two individual branches, the main branch and the side branch, which model the vision-language cross-modality knowledge and query language knowledge, respectively.
The main branch is applied to predict the original query based on the interaction of the information embedded in both the video and masked query, while the side branch is utilized to measure the contribution of the un-masked tokens in masked query for the reconstruction.
By applying a counterfactual cross-modality knowledge in the aggregation of two branches, the unimodal impact of the masked query is extracted.
Because the reconstruction of the masked query should be performed mainly by the cross-modality interaction between the video and masked query, we weaken the contribution of the uni-modal masked query in the reconstruction by directly removing it from the final prediction.

We summarize our major contributions as follows:
(1) We propose a novel method, counterfactual cross-modality reasoning, for weakly supervised video moment localization, aimed at discovering and mitigating potential spurious correlations between different words in the query.
(2) We formulate the query reconstruction task in weakly supervised video moment localization from a causal reasoning perspective, and disentangle the causal effect on the prediction into main-branch and side-branch, which encode the cross-modality and query knowledge respectively.
Through aggregating a counterfactual cross-modality knowledge with the query, the unimodal effect contributed by the masked query is explicitly modeled.
By suppressing this spurious effect, the prediction of the original query is rectified to rely more on the cross-modality knowledge rather than the un-masked words in the query, and thus the efficacy of contrastive learning is directly promoted.
(3) We evaluate the proposed CCR on ActivityNet Captions and Charades-STA benchmark datasets. Experimental results show that our CCR significantly outperforms the state-of-the-art baseline.

\vspace{-0.2cm}
\section{Related Work}
\textbf{Fully supervised video moment localization.}
Video moment localization is formulated in a fully supervised setting in early studies \cite{gao2017tall, zhang2020learning, yuan2019semantic,zhang2020span,li2021proposal}, which is trained based on the exact temporal boundaries of video moment corresponding to each query.
Existing methods can be divided into two categories, which utilize anchor-based and anchor-free paradigms respectively, according to whether they need anchors during training.
Anchor-free methods \cite{zhang2020span,li2021proposal} predict the probability of the temporal boundary for each frame within the given video based on the cross-modality fusion feature in a one-stage manner.
While anchor-based methods \cite{gao2017tall, zhang2020learning, yuan2019semantic} firstly generate a set of video proposals, and then train an interaction model to predict the similarities between these proposals and the given query.
In fully supervised setting, the proposal generation processing can be treated as an additional supervision signal compared to anchor-free methods, and thus the performance of anchor-based methods usually surpasses that of anchor-free methods.
However, it is expensive and sometimes unreliable to manually annotate the precise temporal boundary \cite{lan2023survey}, which indeed restricts the generalization performance of fully supervised methods.

\textbf{Weakly supervised video moment localization.}
To increase the scalability of video moment localization in real life practice, weakly supervised methods \cite{li2017multiple} are introduced with no requirement of the boundary label.
Because there is no precise temporal location of the target moment, most of the existing weakly supervised methods follow anchor-based paradigm to train their models based on the generated proposals.
\cite{chen2020look,huang2021cross,li2017multiple,wang2021weakly} propose to generate video proposals by utilizing sliding temporal windows strategies.
\cite{lin2020weakly} firstly introduces the self-supervised reconstruction of the masked query to connect the information between video and query for weakly supervised video moment localization.
The reconstruction loss is utilized as the similarity measurement between the proposals and query, where the rationale lies in that the visual content inside the matched video proposal should be more helpful comparing to the mismatched ones.
However, generating proposal through a enumerate manner is unreasonable because they are irrelevant with neither video nor query semantic content.
Besides, in order to cover more potential video moment, they have to increase the number of proposals, which cause huge computational cost.
Recent works propose to generate video proposals utilizing learn-based methods.
\cite{cnm,cpl} apply a multi-model transformer to fuse the video and query, which outputs the parameters of a series of temporal weights shaped like Gaussian distributions as the positive proposals and the corresponding negative proposals obtained in a heuristic method.
After that, they feed the temporal proposals along with masked query into the transformer to reconstruct the masked words and apply contrastive learning to contrast between the similarities between the positive and negative proposals.
However, the central step, which is the reconstruction of the masked query, can be turbulent because of the spurious correlation between the masked words and the un-masked ones.
To tackle this issue, we propose Counterfactual Cross-modality Reasoning (CCR) to decouple the causal effect on the prediction of the masked words into the effect of cross-modality fusion knowledge and query knowledge, which are indicated as the main-branch and side-branch respectively, and thus mitigate the spurious correlation inside the query by suppressing the contribution of the side-branch.

\textbf{Causal reasoning in multi-model learning.}
Causal reasoning has shown its capability of resolving the ubiquitous biased training set and spurious correlation issues in multi-model fields including video corpus moment retrieval \cite{yoon2022selective} and video question answering \cite{rubi}.
\cite{cfvqa,rubi} try to reduce the unimodal biases in video question answering through modeling the statistical regularities between the question and the answer, which have similar key idea to our CCR.
However, our CCR is proposed to calibrate the cross-modality contrastive learning process by rectifying the masked query reconstruction, while \cite{cfvqa,rubi} are designed for de-biased answer making.
\cite{zhang2020counterfactual} designs a two-stage paradigm started with generating proposals through coarse contrastive learning between different video-query pairs, and then develops three memory bank based heuristical transformations to apply counterfactual contrastive learning on the generated proposals within a mini-batch to tackle weakly supervised video moment localization.
Nevertheless, heuristically replacing and perturbing the proposal features is not sufficient and reliable for a reasonable counterfactual situation because it heavily relies on the positive and negative proposals generated in the first inter-sample based contrastive learning.
By comparison, our CCR is proposed to facilitate the alignment between fine-grained intra-video proposals and the given query by erasing the spurious correlation hidden in the masked query reconstruction process.

\vspace{-0.3cm}
\section{Method}
\label{sec:method}
\subsection{Preliminary}
Given a natural language query $W$ and a video $V$, video moment localization task aims to localize a video moment $\hat{V}$ corresponding to the semantic information of the query $W$.
To localize this segment, a prediction model $\bm{\eta_\bm{\theta}}(\hat{V}|V, W)$ parameterized by $\bm{\theta}$ is trained to minimize the distance between $\hat{V}$ and the ground truth segment $\tilde{V}$ according to the assessment metric such as intersection-over-union.

However, the ground truth $\tilde{V}$ is unavailable in the weakly supervised setting.
To refine the video-level annotated query to the corresponding target segment level, video proposals are generated as the candidates to semantically match with the query.
To model the semantic alignment between the query and video proposals, the query is partially masked, and the model is trained to learn cross-modality fusion knowledge to reconstruct the original query. This is an efficient methodology to build fine-grained alignment in weakly supervised settings \cite{lin2020weakly}.
Following the scheme illustrated in Figure \ref{fig:intro}, we elaborate on the procedure of query reconstruction-based weakly supervised video moment retrieval as follows:
(1) Given a video $V$ its corresponding query $W$, positive video proposals $\mathcal{S}^{p}=\{S^{p}_i|i=1,...,N^p\}$ and negative video proposals $\mathcal{S}^{n}=\{S^{n}_j|j=1,...,N^n\}$ are obtained by interacting the video and query, where $N_p$ and $N_n$ are the numbers of positive and negative proposals, respectively.
(2) For each proposal $S^{p}_i\in \mathcal{S}^{p}$ and $S^{n}_j \in \mathcal{S}^{n}$, model learns to reconstruct the original query as $\hat{W}^p_i, \hat{W}^n_j$ based on the masked query $\bar{W}$ respectively.
(3) Reconstruction losses $L^p_i=Loss(\hat{W}^p_i,W)$ and $L^n_i=Loss(\hat{W}^n_i,W)$ are utilized to indicate the similarities between the video proposals and the query, where a lower loss implies a higher alignment degree between the proposal and query, and vice versa.
(4) Perform contrastive learning between the reconstruction losses of positive and negative proposals $<\mathcal{S}^p,\mathcal{S}^n>$ to train the model.
(5) During inference, $\mathcal{S}^n$ is neglected, and the reconstruction loss $L^p_i$ for each $S^{p}_i\in \mathcal{S}^p$ is calculated. The proposal with the lowest $L^p_i$ is output as the final localization.
From the scheme above we can find that, query reconstruction is the core step that not only semantically connects vision and language but also serves as the measurement during evaluation.
One of the key factors in establishing semantic alignment between video proposals and queries is to use contrastive learning to increase the discrepancy between the reconstruction losses of positive and negative proposals.
However, the spurious correlation between the masked query and its reconstruction leads to an invalid contrast between positive and negative video proposals, which directly perturbs the cross-modality alignment.

\subsection{Revisit Masked Query Reconstruction in Causality View}
\begin{figure}[t]
\vspace{0.0cm}
  \centering
  \setlength{\abovecaptionskip}{0.0cm}
  \setlength{\belowcaptionskip}{-0.2cm}
  \includegraphics[width=1.0\linewidth]{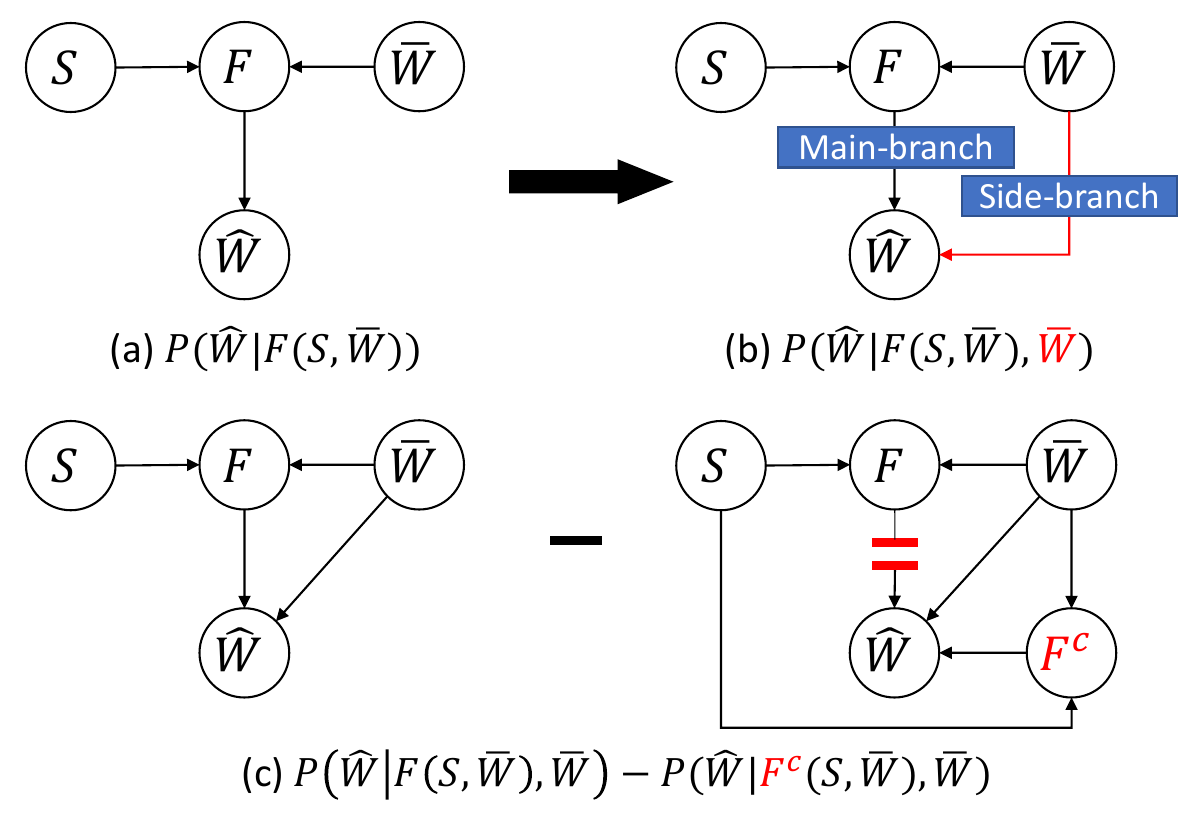}
  \caption{$S$, $\bar{W}$, and $\hat{W}$ indicate the proposal feature, masked query, and reconstructed query, we omit the superscripts $* \in \{p,n\}$ of $S$ in SCM; $F$ is the cross-modality fusion knowledge.
  (a) Conventional query reconstruction SCM based on the cross-modality fusion between video and masked query.
  (b) Modified causal graph decouples the effect on reconstruction into the main-branch and the side-branch, which encode the cross-modal and the unimodal impact of the masked query respectively.
  (c) Removing the spurious correlation between the masked query and its reconstruction by introducing the counterfactual cross-modality knowledge $F^c$.}
  \label{fig:scm}
  \vspace{-0.3cm}
\end{figure}
To model and further mitigate this spurious correlation, we propose to revisit the cross-modality fusion based query reconstruction task from the perspective of the Structured Causal Model (SCM) \cite{glymour2016causal}.
Without loss of generality, we can simplify the notation of each video proposal by omitting all subscripts and just denoting it as $S$.
We introduce the cross-modality knowledge as $F$ and formulate the Structured Causal Model (SCM) of the conventional query reconstruction methodology in Figure \ref{fig:scm} (a), as follows
\begin{equation}\label{eq:scm}
  P(\hat{W}|F(S, \bar{W})),
\end{equation}
where the cross-modality fusion knowledge is the only causal of the query reconstruction, as the $\{S, \bar{W}\}\rightarrow F\rightarrow \hat{W}$ path indicated in Figure \ref{fig:scm} (a).

However, the SCM in Figure \ref{fig:scm} (a) neglects the spurious correlation between the masked query $\bar{W}$ and the final prediction.
Therefore, from a perspective of causality, we modify the conventional SCM by adding a causal connection from $\bar{W}$ directly to $\hat{W}$ to model this spurious correlation, as shown in Figure \ref{fig:scm} (b).
We note that this causal effect is formulated individually from the existing cross-modality fusion knowledge $F$ because the original query can be reconstructed only by the unimodal impact of the masked query, which is induced by certain patterns learned by the model.
Based on the upgraded SCM, we can reformulate the masked query reconstruction as the result of the aggregated effect of two branches
\begin{equation}\label{eq:updated-scm}
  P(\hat{W}|F(S, \bar{W}), \bar{W}),
\end{equation}
where $F(S, \bar{W})$ and $\bar{W}$ indicate the causal effects of main-branch $F\rightarrow \hat{W}$ and side-branch $\bar{W}\rightarrow \hat{W}$ respectively.
To explicitly model the spurious correlation implying in the side-branch $\bar{W}\rightarrow \hat{W}$, we cut off the potential impact from the main-branch $F\rightarrow \hat{W}$ by applying a counterfactual cross-modality knowledge $F^c$ which does not provide any useful information for establishing semantic interaction between the video and query, and hence, does not assist in the reconstruction of the original query.
As illustrated in Figure \ref{fig:scm} (c), by replacing the cross-modality effect $F(S, \bar{W})$ in Equation (\ref{eq:updated-scm}) by $F^c(S, \bar{W})$, we obtain the counterfactual reconstruction as
 \begin{equation}\label{eq:cf-scm}
  P(\hat{W}|F^c(S, \bar{W}), \bar{W}),
\end{equation}
which is known as the total indirect effect in causality \cite{rubin1978bayesian}.
Following this, we can finally eliminate the spurious correlation between the masked query and original query reconstruction as
 \begin{equation}\label{eq:tie}
  P(\hat{W}|F(S, \bar{W}), \bar{W}) - P(\hat{W}|F^c(S, \bar{W}), \bar{W}),
\end{equation}
as shown in Figure \ref{fig:scm} (c), where the effect on reconstruction is attributed solely to the interaction $F$ between video proposal and query.
Therefore, our proposed methodology can effectively capture the true causal relationship between the reconstruction and cross-modality knowledge.

It is important to note that reducing the spurious correlation induced by masked query solely by reconstructing queries from the video proposal is not feasible.
This is due to the presence of redundant visual content in video proposal, and masked query plays a crucial attentional role in establishing cross-modality alignments.

\subsection{Counterfactual Cross-modality Reasoning}
\begin{figure*}[t]
\vspace{0.0cm}
  \setlength{\abovecaptionskip}{0.0cm}
  \setlength{\belowcaptionskip}{-0.2cm}
  \centering
  \centerline{\includegraphics[width=1\linewidth]{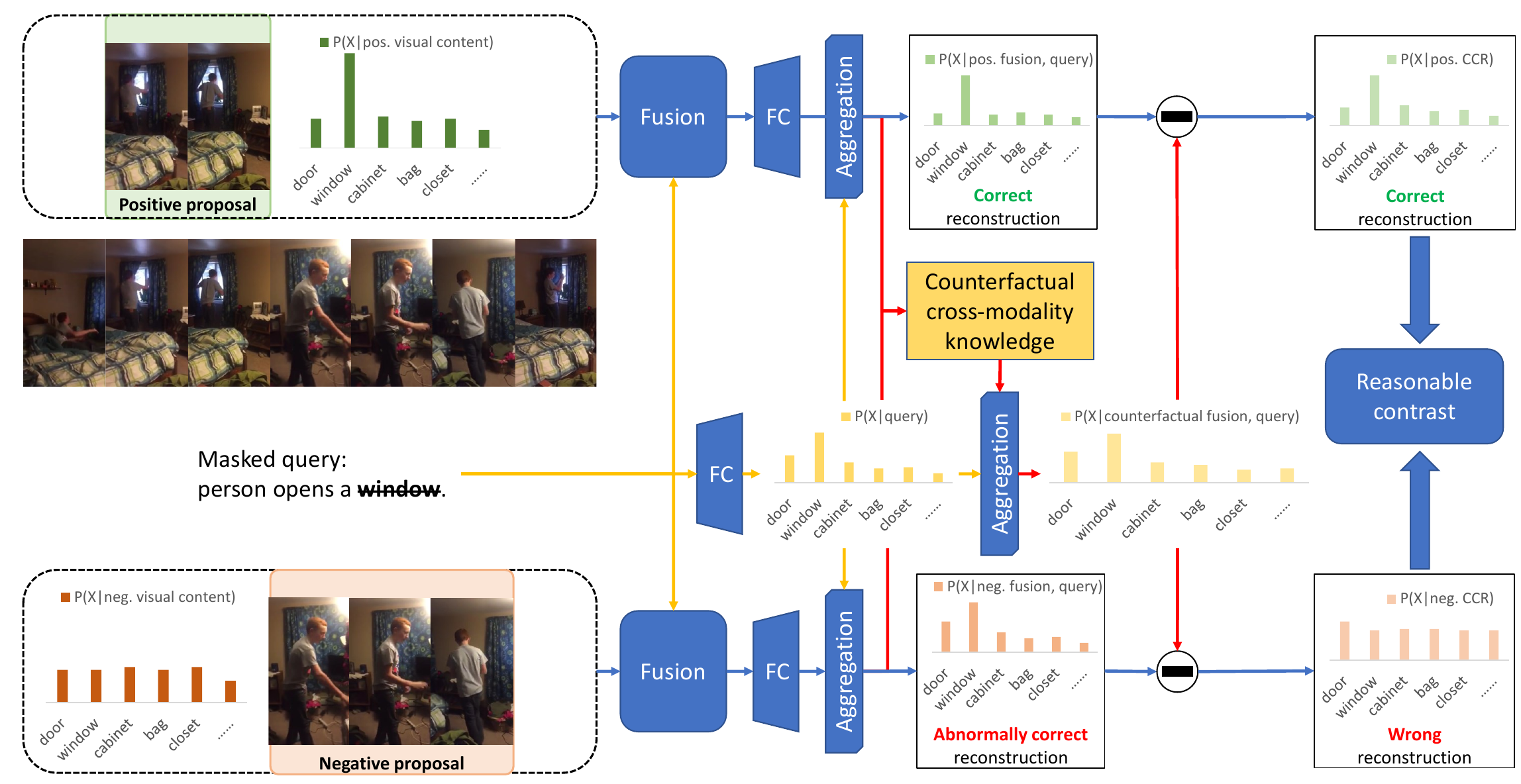}}
  \caption{Overview of our proposed Counterfactual Cross-modality Reasoning (CCR) scheme.
  The main-branches associated with the positive and negative proposals, which are indicated in the blue connections, reconstruct the original query correctly and abnormally correctly, respectively. The reason behind the abnormal reconstruction is the spurious correlation between the masked query and its reconstruction.
  \emph{e.g.} because of the biased distribution of the co-occurrence of words ``open'' and ``person'', the masked ``window'' can be easily reconstructed only based on the certain pattern of query even though ``window'' is not correlated with the visual content in the negative proposal $P(X|neg. \  visual \  content)$.
  This spurious correlation between the masked words ``window'' and the remaining ones, which is noted as the side-branch in the yellow connections, is modeled as $P(X|query)$, and is aggregated with the counterfactual cross-modality knowledge to obtain the total effect of the masked query $P(X|counterfactual \  fusion, query)$.
  Finally, by suppressing this spurious correlation in both the reconstructions of positive and negative proposals, a reasonable contrast can be applied between the rectified prediction $P(X|pos. \ CCR)$ and $P(X|neg. \ CCR)$.
  }
  \label{fig:fw}
  \vspace{-0.3cm}
\end{figure*}
In this subsection, we present a detailed description of the proposed Counterfactual Cross-modality Reasoning (CCR), as illustrated in Figure \ref{fig:fw}.
The central idea of CCR is to mitigate the spurious correlation between the masked query and its reconstruction.
To achieve this, we propose to decouple the total effect of masked query reconstruction as main-branch and side-branch respectively, and reconstruct the original query by combining these two branches.
Given video proposal feature $S$ and masked query feature $\bar{q}$, we model the main-branch in Equation (\ref{eq:scm}) as
\begin{equation}\label{eq:mainbranch}
  P(\hat{W}|F(S, \bar{W})): \ \phi_q = \pi(\chi(S,\bar{q})),
\end{equation}
where $\phi_q$ is the reconstruction logit produced by cross-modality knowledge, $\chi(\cdot)$ indicates a cross-modality interaction module, and $\pi(\cdot)$ is a fully connected layer, which projects the fusion feature from latent space to word embedding space.
Meanwhile, we model the side-branch highlighted in Figure \ref{fig:scm} (b) as the impact of the masked query on the reconstruction as
\begin{equation}\label{eq:sidebranch}
  P(\hat{W}|\bar{W}): \ \psi_q = \pi(\bar{q}),
\end{equation}
where $\psi_q$ denotes the reconstruction logit only generated by the masked query $\bar{q}$.
Hence we can reconstruct the original query by combining the prediction logits of these two branches as
\vspace{-0.1cm}
\begin{equation}\label{eq:recon}
\begin{aligned}
  P(\hat{W}|F(S, \bar{W}), \bar{W}): \ \hat{q} &= \rho(\phi_q, \psi_q)  \\
  &= \rho\big(\pi(\chi(S,\bar{q})), \pi(\bar{q})\big),
  \end{aligned}
  \vspace{-0.05cm}
\end{equation}
where $\hat{q}$ is the final reconstruction logit, and $\rho(\cdot)$ is an aggregation function.

To better isolate the impact of the side-branch on original query reconstruction, we propose a counterfactual approach that cuts off the effect of the main-branch.
This ensures that the cross-modality knowledge contributes nothing to the reconstruction of the original query, allowing us to focus solely on the contribution of the side-branch.
To create a counterfactual main-branch, we force the cross-modality knowledge to predict the original query randomly, regardless of the input video proposal and masked query.
As a result, the final reconstruction obtained by aggregating the predictions generated by cross-modality knowledge and masked query will solely rely on the latter.
To achieve this, we modify the cross-modality prediction $\phi_q$ in Equation (\ref{eq:recon}) into a uniform logit in this counterfactual situation, which is parameterized by a learnable scalar $\mu$ \cite{cfvqa}.
By aggregating the logits produced by $\mu$ and masked query, we obtain the counterfactual reconstruction logit $\hat{q}^c$ as
\vspace{-0.2cm}
\begin{equation}\label{eq:cf-recon}
  P(\hat{W}|F^c(S, \bar{W}), \bar{W}): \ \hat{q}^c = \rho(\mu, \pi(\bar{q})),
\vspace{-0.2cm}
\end{equation}
where $\mu$ solely impacts the absolute value of the logit $\hat{q}^c$, but it does not alter the relative value that represents the final reconstruction.

Based on the reconstruction $\hat{q}$ and its counterfactual version $\hat{q}^c$, which is altered to rely solely on the masked query by incorporating counterfactual cross-modality knowledge into the reconstruction process, we propose to mitigate the spurious correlation between the masked query and its reconstruction by removing the unimodal effect as
\begin{equation}\label{eq:final}
\begin{aligned}
  \hat{W} &= Softmax(\hat{q}-\hat{q}^c) \\
  &= Softmax\Big( \rho\big(\pi(\chi(S,\bar{q})), \pi(\bar{q})\big)-\rho(\mu, \pi(\bar{q})) \Big).
  \end{aligned}
\end{equation}
To prevent a trivial solution of Equation (\ref{eq:final}), we use a non-linear aggregation function
\vspace{-0.15cm}
\begin{equation}\label{eq:fusion}
  \rho(x, y)=x\odot Sigmoid(y)
\end{equation}
\vspace{-0.0cm}
to combine the effects of the main-branch and side-branch.

\vspace{-0.2cm}
\subsection{Training and inference}
\vspace{-0.1cm}

We embed the proposed CCR into an off-the-shelf query reconstruction based contrastive learning scheme \cite{cpl}.
To begin with, a multi-modal transformer is applied to interact the video with the query as the implementation of $\chi(\cdot)$.
Then, for each video-query pair, the transformer generates $n$ positive proposal $\mathcal{S}^p=\{S^p_i|i=1,...,N^p\}$ and their corresponding $2n$ intra-video negative proposals $\mathcal{S}^n=\{S^{n_k}_{j}|j=1...N^p,k=1,2\}$, and the whole video is treated as the reference proposal $S^r$.
The diversity of the positive proposals is ensured by a penalization term \cite{lin2017structured}
\begin{equation}\label{eq:lossdiv}
  \ell_{div} = \| \Omega \Omega^{\top} - \lambda I\|_F^2,
\end{equation}
where $\lambda$ is a hyperparameter, and $\Omega = cat[\omega^p_1;...;\omega^p_{N^p}]$ where $\omega^p_i$ is the temporal weight of $S^p_i$ \cite{cpl}.

For each triplet $<S^p_i, S^{n_1}_{i}, S^{n_2}_{i}>$ and $S^r$, by omitting the subscript $i$, the reconstruction losses of positive, negative and reference proposals, which are denoted as $\ell^p_c$, $\ell^r_c$, $\ell^{n_1}_{c}$, and $\ell^{n_2}_{c}$ respectively, are rectified through our proposed CCR to minimize the losses of the spurious correlation mitigated reconstruction $\hat{W}$ in Equation (\ref{eq:final}) and the aggregated logit $\hat{q}$ in Equation (\ref{eq:recon}) corresponding to all the proposals as
\begin{equation}\label{eq:recloss}
\begin{aligned}
  \ell^*_{c} &= CE(\hat{W}^*,W) + CE(Softmax(\hat{q}^*),W) \\
  * & \in \{p,r,n_1,n_2\},
  \end{aligned}
\end{equation}
where $CE(\cdot)$ is the cross entropy loss.
Thus, the counterfactual intra-video contrastive loss \cite{cpl} for each video-query pair is obtained as
\begin{equation}\label{eq:ctloss}
\begin{aligned}
  \ell_c =& max(0, \alpha_p + \ell^p_c - \ell^r_c) \\
  &+ max(0, \alpha_n + \ell^p_c - \ell^{n_1}_{c}) + max(0, \alpha_n + \ell^p_c - \ell^{n_2}_{c}),
  \end{aligned}
\end{equation}
where $\alpha_p$ and $\alpha_n$ are hyperparameters.

Meanwhile, we train the fully connected layer $\pi$ to minimize the reconstruction error given the masked query as
\begin{equation}\label{eq:lossq}
  \ell_{q} = CE(Softmax(\psi_q),W). 
\end{equation}
Then we optimize the cross-modality module $\chi$ and projection layer $\pi$ with respect to
\begin{equation}\label{eq:loss}
  \ell = \ell_c + \ell_{q} + \ell_{div}.
\end{equation}
Additionally, $\mu$, which provides the counterfactual cross-modality knowledge, is optimized individually from $\chi$ and $\pi$ as
\begin{equation}\label{eq:klloss}
\begin{aligned}
  \ell^*_{kl} &= KL(Softmax(\hat{q}^*)|Softmax(\hat{q}^c)), * \in \{p,r,n_1,n_2\}\\
  \ell_{kl} &= \sum_{*\in \{p,r,n_1,n_2\}}\ell^*_{kl},
  \end{aligned}
\end{equation}
for all the proposals to minimize the Kullback-Leible divergence between $\hat{q}^*$ and its counterfactual version $\hat{q}^c$ to prevent the rectified reconstruction from being dominated by one of them \cite{cfvqa}.

During inference, given an untrimmed video $V$ and its corresponding query $W$, the multi-modal fusion module $\chi(\cdot)$ first encode them to generate the set of positive proposals $\mathcal{S}^p=\{S^p_i|i=1,...,N^p\}$.
Following the experiment setting in \cite{cpl}, a vote-based strategy \cite{zhou2021ensemble} is utilized to select the best proposal as the output.
More specifically, for each positive proposal, we compute its Intersection over Union (IoU) with the other $N_p-1$ positive proposals, and the sum of IoUs represents the number of votes it receives.
Ultimately, we select the positive proposal with the highest number of votes as the final prediction.
The overall training and inference procedures are presented in Algorithm \ref{alg:ccr}.
\begin{algorithm}[t]
\caption{Counterfactual Cross-modality Reansoning (CCR) for each video-query pair in dataset}\label{alg:ccr}
\KwData{positive proposal features $\mathcal{S}^{p}$, negative proposal features $\mathcal{S}^{n}$, masked query $\bar{q}$, query $W$, weight matrix of positive proposals $\Omega$}
\KwResult{cross-modality fusion module $\chi$, prediction layer $\pi$, uniform logit $\mu$}

\While{$S^p \in \mathcal{S}^{p}$ and $S^{n_1}$, $S^{n_2} \in \mathcal{S}^{n}$ }
{
$ \psi_q \leftarrow \pi(\bar{q})$\;
$\hat{q}^c \leftarrow \rho(\mu, \psi_q)$\;
\For{$* \in \{p,n_1,n_2,r\}$}{
calculate $\phi_q^*,\hat{q}^*,\hat{W}^*$ \emph{w.r.t.} Equation (\ref{eq:mainbranch}), (\ref{eq:sidebranch}), and (\ref{eq:final})\;
}

\eIf{training}{
$\ell_{div} \leftarrow \| \Omega \Omega^{\top} - \lambda I\|_F^2$\;
$\ell_{q} \leftarrow CE(Softmax(\psi_q),W)$\;
calculate $\ell_c$ \emph{w.r.t.} Equation (\ref{eq:recloss}) and (\ref{eq:ctloss})\;
update $\chi,\pi$ \emph{w.r.t.} $\ell_{q}$, $\ell_{div}$ and $\ell_c$\;
calculate $\ell_{kl}$ \emph{w.r.t.} Equation (\ref{eq:klloss})\;
update $\mu$ \emph{w.r.t.} $\ell_{kl}$\;
}
{
return the best proposal in $\mathcal{S}^p$ according to vote-strategy \cite{cpl}\;
}
}
\end{algorithm}
\vspace{1.4cm}

\renewcommand{\arraystretch}{1.0} 
\begin{table}[t]
  \centering
  \vspace{0.2cm}
  \setlength{\abovecaptionskip}{-0.0cm}
  \setlength{\belowcaptionskip}{-0.2cm}
  \caption{Comparison of $mIoU$ on Charades-STA and ActivityNet Captions datasets.
  The best result for each metric is displayed in bold, and the second-best result is marked in red. CPL-R denotes the result reproduced based on the official CPL repository and is used to replace the original CPL result in the top two rankings.}
  \label{tab:miou}
  \setlength{\tabcolsep}{0.9mm}{
    \begin{tabular}{ccccc}
    \toprule
    \multirow{2.5}{*}{Methods}&
    \multicolumn{2}{c}{Charades-STA}&\multicolumn{2}{c}{ActivityNet Captions}\cr
    \cmidrule(lr){2-3} \cmidrule(lr){4-5}
    & $R@1,mIoU$ & $R@5,mIoU$ & $R@1,mIoU$ & $R@5,mIoU$\\
    \midrule
    WS-DEC \cite{duan2018weakly} & - & - & 28.23 & - \\
    CTF \cite{chen2020look} & 27.3 & - & 32.20 & - \\
    WSLLN \cite{gao2020weakly} & - & - & 32.20 & - \\
    WSRA \cite{fang2020weak}& 31.00 & - & -&- \\
    VCA \cite{wang2021visual} & 38.49 & - & 33.15&- \\
    LCNet \cite{yang2021local}& 38.94 & - & 34.29&- \\
     CPL \cite{cpl} & 43.48 & - & -  & - \\
     CPL-R\cite{cplrepo} & \textcolor[rgb]{1.00,0.00,0.00}{43.50} & \textcolor[rgb]{1.00,0.00,0.00}{67.70} & \textcolor[rgb]{1.00,0.00,0.00}{35.71}  & \textcolor[rgb]{1.00,0.00,0.00}{43.78} \\
    \midrule
      Ours & \textbf{44.66} & \textbf{67.86} & \textbf{36.69}  & \textbf{53.37} \\

    \bottomrule
    \end{tabular}
    }
    \vspace{-0.6cm}
\end{table}
\vspace{-1.0cm}
\section{Experiments}
\vspace{-0.1cm}

\subsection{Implementation Details}
\vspace{-0.1cm}
In this paper, we measure the effectiveness of moment localization using temporal Intersection over Union ($IoU$), which is the ratio between the temporal overlap and union of the segment predicted by the model and the ground truth moment.
Specifically, we use ``$R@a,mIoU$'' to evaluate localization performance, which is the average $IoU$ of the $a$ predictions with the lowest reconstruction loss based on the rectified reconstruction $\hat{q}^p - \hat{q}^c$.
Additionally, we use ``$R@a,IoU=b$'' as an evaluation metric to further assess performance, which means there is at least one predicted moment with a temporal $IoU$ larger than $b$ among the top $a$ predictions.
We reproduce the CPL \cite{cpl} as our baseline on one NVIDIA GeForce RTX 3090 GPU, and follow all the hyperparameter settings provided by their official repository \cite{cplrepo} to ensure a fair comparison.

\subsection{Datasets}
\vspace{-0.1cm}
We evaluate our proposed CCR on two widely used benchmark datasets, Charades-STA \cite{sigurdsson2016hollywood} and ActivityNet-Captions \cite{caba2015activitynet}.

\textbf{Charades-STA.}
The Charades-STA dataset \cite{sigurdsson2016hollywood} consists of 16,128 video-query pairs generated by 6,672 videos, with an average video duration of 29.96 seconds.
Following \cite{cpl}, we trained our model on the training set, which contains 12,408 video-query pairs, and evaluated the performance on the test set, which contains 3,720 video-query pairs.

\textbf{ActivityNet-Captions.}
The ActivityNet-Captions dataset \cite{caba2015activitynet} contains 19,209 videos with an average duration of 117.6 seconds.
Following \cite{cpl}, we split the dataset into training, validation, and test sets, which contain 37,417, 17,505, and 17,031 video-query pairs, respectively.
\vspace{-0.0cm}

\vspace{-0.2cm}
\subsection{Comparison with state-of-the-arts}
\vspace{-0.1cm}
\renewcommand{\arraystretch}{1.0} 
\begin{table}[t]
  \centering
  \vspace{0.2cm}
  \setlength{\abovecaptionskip}{-0.0cm}
  \setlength{\belowcaptionskip}{-0.2cm}
  \caption{Performance Comparison on Charades-STA. The best result for each metric is displayed in bold, and the second-best result is marked in red. CPL-R denotes the result reproduced based on the official CPL repository and is used to replace the original CPL result in the top two rankings. The results of CRM (indicated as $CRM^\dagger$) are not included in the top two rankings because it needs paragraph-video annotations during training.}
  \label{tab:charades}
  \setlength{\tabcolsep}{1.7mm}{
    \begin{tabular}{ccccccc}
    \toprule
    \multirow{2.5}{*}{Methods}&
    \multicolumn{3}{c}{$R@1,IoU=$}&\multicolumn{3}{c}{$R@5,IoU=$}\cr
    \cmidrule(lr){2-4} \cmidrule(lr){5-7}
    & 0.3 & 0.5 & 0.7 & 0.3 & 0.5 & 0.7 \\
    \midrule
      TGA \cite{mithun2019weakly} & 32.14 & 19.94 & 8.84 & 86.58	&65.52	&33.51 \\
      SCN \cite{lin2020weakly}	&42.96	&23.58	&9.97 &95.56	&71.8	&38.87 \\
      CTF \cite{chen2020look} & 39.8	&27.3	&12.9  & - & - &-\\
    WSTAN \cite{wang2021weakly}	&43.39	&29.35	&12.28	&93.04	&76.13	&41.53 \\
    BAR \cite{wu2020reinforcement}&44.97	&27.04	&12.23 & -& -&-\\
    WSRA \cite{fang2020weak}& 50.13 &31.20 &11.01 &86.75 &70.50 &39.02 \\
    VLANet \cite{ma2020vlanet} &45.24	&31.83	&14.17	&95.7	&82.85	&33.09 \\
    LoGAN \cite{tan2021logan}& 48.04 & 31.74 & 13.71 & 89.01 & 72.17 & 37.58 \\
    MARN \cite{song2020weakly}& 48.55 &31.94 &14.81 &90.70 &70.00 &37.40\\
    CCL \cite{zhang2020counterfactual} & - &33.21 &15.68& - &73.50& 41.87 \\
    $CRM^\dagger$\cite{huang2021cross} &53.66& 34.76& 16.37& -& -& - \\
     CNM\cite{cnm} &60.39 &35.43 &15.45& - &- &-\\
    LCNet\cite{yang2021local}& 59.60& 39.19& 18.87 &94.78 &80.56 &45.24 \\
    RTBPN\cite{zhang2020regularized} &60.04 &32.36& 13.24& \textcolor[rgb]{1.00,0.00,0.00}{97.48} & 71.85& 41.18 \\
    VCA\cite{wang2021visual}&58.58& 38.13& 19.57& \textbf{98.08}& 78.75& 37.75 \\
        CPL\cite{cpl} & 65.99 & 49.05 & 22.61  & 96.99 & 84.71 & 52.37 \\
      CPL-R\cite{cplrepo} & \textcolor[rgb]{1.00,0.00,0.00}{66.53} & \textcolor[rgb]{1.00,0.00,0.00}{49.43} & \textcolor[rgb]{1.00,0.00,0.00}{22.36}  & 96.80 & \textcolor[rgb]{1.00,0.00,0.00}{84.20} & \textcolor[rgb]{1.00,0.00,0.00}{52.18} \\
      \midrule
      Ours & \textbf{68.59} & \textbf{50.79} & \textbf{23.75}  & 96.85 & \textbf{84.48} & \textbf{52.44}\\

    \bottomrule
    \end{tabular}
    }
    \vspace{-0.7cm}
\end{table}
We compare the performance of our proposed CCR with state-of-the-art methods on Charades-STA and ActivityNet Captions using $R@a,mIoU$ and $R@a,IoU=b$, where $a\in\{1,2\}$ and $b\in\{0.1,0.3,0.5,0.7\}$.
The results are presented in Table \ref{tab:miou}, Table \ref{tab:charades}, and Table \ref{tab:acnet}, respectively.
Because there are significant differences between the performances of our reproduction and CPL \cite{cpl}, we additionally include the results of our reproduction as CPL-R for comparison.
Directly comparing CRM \cite{huang2021cross} with other methods, including ours, is unfair because CRM requires multiple queries that appear sequentially in the video for training, and hence we have not highlighted its results in the top two rankings.

We compare the average temporal $IoU$ between our proposed CCR and the existing methods in Table \ref{tab:miou}.
Our proposed CCR outperforms the current state-of-the-art method CPL in all evaluation metrics, demonstrating a remarkable improvement of over $9\%$ in the $R@5,mIoU$ metric on the ActivityNet Captions dataset.
In Table \ref{tab:charades}, our CCR overall surpasses both CPL and CPL-R according to $R@1$ metrics, with an average absolute gain of about $2\%$.
For $R@5$ metrics, our CCR outperformed CPL overall.
On ActivityNet Captions dataset, we surpass the baseline and outperform it by approximately $3\%$ and $9\%$ on average for $R@1$ and $R@5$ metrics, respectively.
The significant gain achieved by CCR on the ActivityNet Captions dataset is due to the more variational visual content, which increases the frequency of negative proposals producing abnormal correct reconstructions as illustrated in Figure \ref{fig:intro} and Figure \ref{fig:fw}.
Our CCR is designed to mitigate this issue, leading to its superior performance on this dataset.
Our CCR achieves comparable performance to other methods, and on average performs better than LCNet and VCA on $mIoU$ in Table \ref{tab:miou}, despite their state-of-the-art performance on $R@5, IoU=0.3,0.5$ metrics.

\begin{table}[t]
  \centering
  \vspace{0.0cm}
  \setlength{\abovecaptionskip}{-0.0cm}
  \setlength{\belowcaptionskip}{-0.2cm}
  \caption{Performance comparison on ActivityNet-Captions.
  The best result for each metric is displayed in bold, and the second-best result is marked in red. CPL-R denotes the result reproduced based on the official CPL repository and is used to replace the original CPL result in the top two rankings. The results of CRM (indicated as $CRM^\dagger$) are not included in the top two rankings because it needs paragraph-video annotations during training.}
  \label{tab:acnet}
  \setlength{\tabcolsep}{1.7mm}{
    \begin{tabular}{ccccccc}
    \toprule
    \multirow{2.5}{*}{Methods}&
    \multicolumn{3}{c}{$R@1,IoU=$}&\multicolumn{3}{c}{$R@5,IoU=$}\cr
    \cmidrule(lr){2-4} \cmidrule(lr){5-7}
    & 0.1 & 0.3 & 0.5 & 0.1 & 0.3 & 0.5 \\
    \midrule
    CTF \cite{chen2020look} & 74.2& 44.3 &23.6 &- &- &- \\
    EC-SL \cite{chen2021towards}& 68.48 &44.29 &24.16 &- &- &- \\
    MARN \cite{song2020weakly}& - &47.01 &29.95 &- &72.02 &57.49 \\
    SCN \cite{lin2020weakly}& 71.48 &47.23 &29.22 &90.88 &71.56 &55.69 \\
    BAR \cite{wu2020reinforcement}& - &49.03 &30.73 &- &- &- \\
    RTBPN \cite{zhang2020regularized}& 73.73 &49.77& 29.63 &\textcolor[rgb]{1.00,0.00,0.00}{93.89} &\textcolor[rgb]{1.00,0.00,0.00}{79.89}& 60.56 \\
    CCL\cite{zhang2020counterfactual} & - &50.12 &\textcolor[rgb]{1.00,0.00,0.00}{31.07} &- &77.36 &61.29 \\
    WSTAN\cite{wang2021weakly} & \textcolor[rgb]{1.00,0.00,0.00}{79.78} &52.45& 30.01 &93.15 &79.38 &\textbf{63.42} \\
    $CRM^\dagger$\cite{huang2021cross} & 81.61 &55.26 &32.19 &- &- &- \\
    CNM\cite{cnm} & 78.13 &\textbf{55.68}& \textbf{33.33} &- &- &- \\
     VCA \cite{wang2021visual} & 67.96 &50.45 &31.00 &92.14 &71.79 &53.83 \\
    LCNet\cite{yang2021local} & 78.58 &48.49 &26.33 &\textbf{93.95} &\textbf{82.51} &\textcolor[rgb]{1.00,0.00,0.00}{62.66} \\
      CPL\cite{cpl} & 82.55 & 55.73 & 31.37  & 87.24 & 63.05 & 43.13 \\
      CPL-R\cite{cplrepo} & 78.13 & 51.19 & 28.19  & 88.23 & 62.16 & 40.04 \\
      \midrule
      Ours & \textbf{80.32} & \textcolor[rgb]{1.00,0.00,0.00}{53.21} & 30.39  & 91.44 & 71.97 & 56.50\\
    \bottomrule
    \end{tabular}
    }
    \vspace{-0.5cm}
\end{table}
\vspace{-0.2cm}
\subsection{Ablation Studies}
\vspace{-0.1cm}
\textbf{Generation of counterfactual cross-modality knowledge.}
The only additional parameter in our proposed method compared to the baseline is the counterfactual cross-modality knowledge $\mu$.
In addition to the uniform distribution presented in Section \ref{sec:method}, we also explore two other possible generation methods for $U$, referred to as ``Average'' and ``Random selected'', and evaluate their effectiveness on the Charades-STA dataset.
As illustrated in Table \ref{tab:abl-u}, our experiments reveal that the replacement within mini-batch is insufficient to generate counterfactual cross-modality knowledge in this scenario.
In this situation, the uniform prediction strategy outperforms the other two methods for generating $\mu$.

\textbf{Aggregation of main-branch and side-branch.}
As discussed in Section \ref{sec:method}, we non-linearly aggregate the effects of main-branch and side-branch as
\begin{equation}\label{eq:abl-fusion}
  \rho(\phi_q, \psi_q)=\phi_q \odot Sigmoid(\psi_q),
\end{equation}
which allows us to obtain the total impact on query reconstruction.
We also implemented $\rho$ as another heuristic non-linear summation and learnable projection network to evaluate their corresponding performance, as presented in Table \ref{tab:abl-rho}.
However, both the learning-based and non-linear summation methods achieved lower performance compared to the aggregation applied in Equation (\ref{eq:abl-fusion}).

\renewcommand{\arraystretch}{1.3} 
\begin{table}[t]
  \centering
  \vspace{0.0cm}
  \setlength{\abovecaptionskip}{-0.0cm}
  \setlength{\belowcaptionskip}{-0.0cm}
  \caption{Ablation study on different manners of generating counterfactual cross-modality knowledge.
  The best result is indicated in bold.
  ``Average'' means that $\mu$ is set as the average prediction of main-branch in a mini-batch, and ``Random selected'' denotes that $\mu$ is randomly selected from the predictions within a mini-batch.}
  \label{tab:abl-rho}
  \setlength{\tabcolsep}{1.5mm}{
    \begin{tabular}{ccccc}
    \toprule
    \multirow{2}{*}{\makecell[c]{Counterfactual\\ cross-modality \\knowledge $\mu$}}&
    \multicolumn{4}{c}{$R@1$}\cr
    \cmidrule(lr){2-5}
     & $mIoU$ & $IoU=0.3$ & $IoU=0.5$ & $IoU=0.7$ \\
    \midrule
      Baseline & 43.50 & 66.53 & 49.43 & 22.36 \\
      Average & 43.69& 66.14 & 49.02 & 22.95 \\
      Random selected &44.01& 67.51 & 49.87 & 23.02 \\
      Uniform & \textbf{44.66} & \textbf{68.59} & \textbf{50.79} & \textbf{23.75} \\
    \bottomrule
    \end{tabular}
    }
    \vspace{-0.5cm}
\end{table}

\renewcommand{\arraystretch}{1.0} 
\begin{table}[t]
  \centering
  \vspace{0.0cm}
  \setlength{\abovecaptionskip}{-0.0cm}
  \setlength{\belowcaptionskip}{-0.2cm}
  \caption{Ablation study on different aggregation manners of the main-branch and side-branch.
  The best result is indicated in bold.
  $Proj(cat[x;y])$ presents we first concatenate $x$ and $y$ along the embedding dimension, then utilize a learnable linear layer to project it back to the original space.
  }
  \label{tab:abl-u}
  \setlength{\tabcolsep}{1.0mm}{
    \begin{tabular}{ccccc}
    \toprule
     \multirow{2.5}{*}{\makecell[c]{Branch aggregation\\ $\rho(x,y)$}}&
    \multicolumn{4}{c}{$R@1$}\cr
    \cmidrule(lr){2-5}
     & $mIoU$ & $IoU=0.3$ & $IoU=0.5$ & $IoU=0.7$ \\
    \midrule
      Baseline &43.50& 66.53 & 49.43 & 22.36 \\
      $Proj(cat[x;y])$ &44.30& 67.89 & 50.76 & 23.65 \\
      $Sigmoid(x+y)$ &44.21& 68.13 & 49.22 & 22.37 \\
      $x \odot Sigmoid(y)$ & \textbf{44.66} & \textbf{68.59} & \textbf{50.79} & \textbf{23.75} \\
    \bottomrule
    \end{tabular}
    }
    \vspace{-0.2cm}
\end{table}

\begin{figure}[t]
\vspace{-0.2cm}
  \centering
  \setlength{\abovecaptionskip}{-0.1cm}
  \setlength{\belowcaptionskip}{-0.2cm}
  \includegraphics[width=1.0\linewidth]{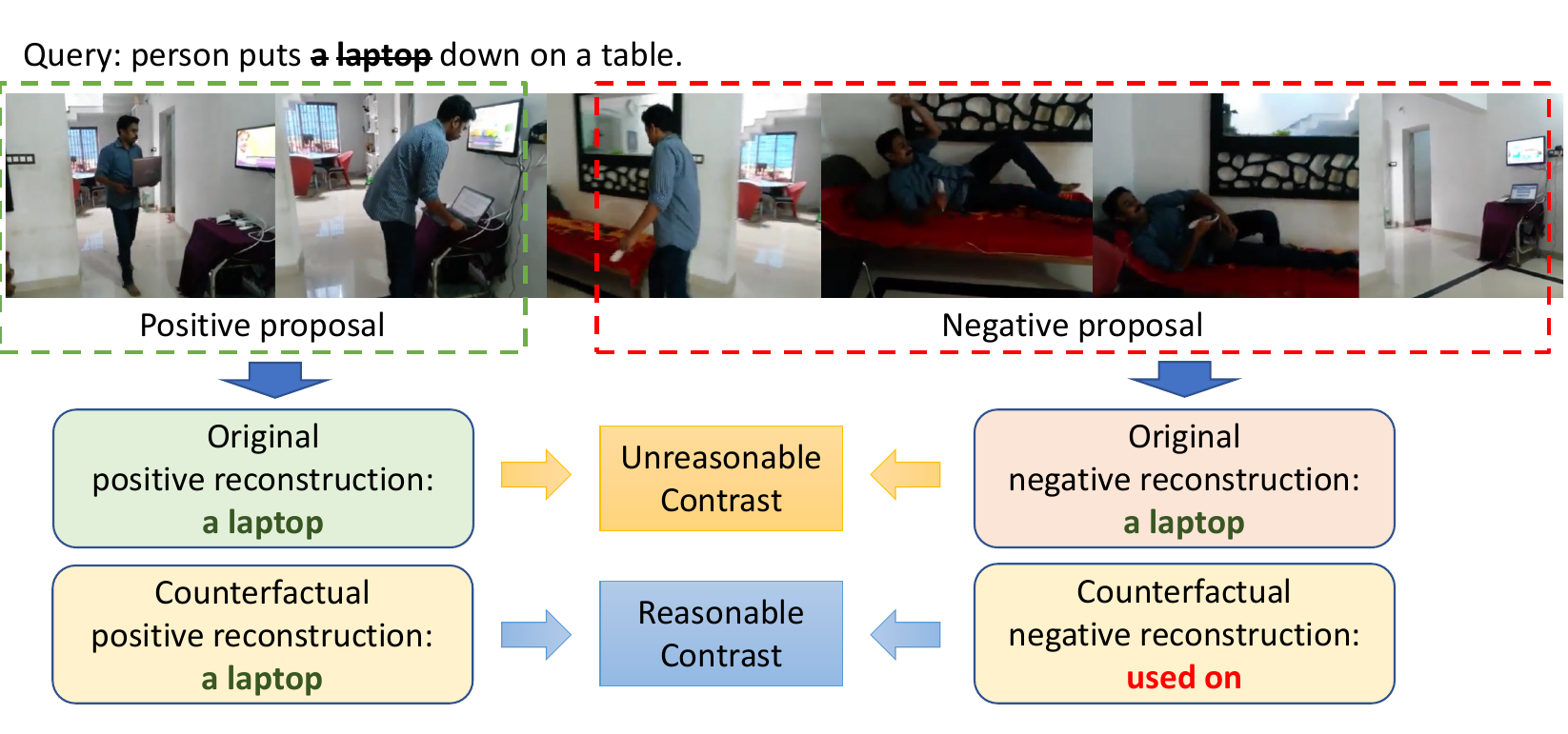}
  \caption{A qualitative example that highlights the contrasts between reconstructions with and without CCR.
  Even though there's no visual content pertaining to the ``laptop'' masked in the query of the negative proposal, the words ``a laptop'' were still successfully reconstructed, rendering the contrast between it and the positive proposal invalid. Utilizing our proposed CCR, the reconstructions are rectified to produce correct and incorrect outcomes for positive and negative proposals, respectively. }
  \label{fig:vis}
  \vspace{-0.3cm}
\end{figure}

\textbf{Qualitative results.}
We provide a qualitative example in Figure \ref{fig:vis} to further illustrate the effectiveness of our proposed CCR.
In this video, a person first places a laptop on a table, as described in the query, then lies on a sofa, and finally watches TV while holding a remote.
The original reconstruction generated by the positive proposal is the same as that of negative proposals, which includes ``a laptop''.
Therefore, the contrast between positive and negative proposals is invalid.
Our proposed CCR rectifies the reconstructions to predict the wrong answer ``used on'' for the masked words, enabling reasonable contrastive learning and further improving the cross-modality alignment.

\vspace{-0.3cm}
\section{Conclusion}
\vspace{-0.15cm}
In this paper, we introduce a novel Counterfactual Cross-modality Reasoning (CCR) method, which addresses the challenge of weakly supervised video moment localization.
We focus on the problem of unreasonable contrastive learning, which arises due to the spurious correlation between masked and unmasked query words.
This issue is commonly overlooked by current state-of-the-art query reconstruction based methods.

To overcome this problem, we first model the impact on query reconstruction as a combination of cross-modality driven main-branch and query-driven side-branch. We then extract the spurious correlation induced by the unimodal impact by applying counterfactual cross-modality knowledge during the aggregation process. Finally, we address the problem of spurious correlation by removing it from the reconstructions of positive and negative proposals, enabling reasonable contrastive learning.
\vspace{-0.3cm}
\section*{Acknowledgments}
\vspace{-0.15cm}
This work was supported in part by the National Natural Science Foundation of China No. 61976206 and No. 61832017, Beijing Outstanding Young Scientist Program NO. BJJWZYJH012019100020098, Beijing Academy of Artificial Intelligence (BAAI), the Fundamental Research Funds for the Central Universities, the Research Funds of Renmin University of China 21XNLG05, and Public Computing Cloud, Renmin University of China.

\bibliographystyle{ACM-Reference-Format}
\balance
\bibliography{sample-base}

%
%
%
%
%
%
%
%

\end{document}